\documentclass[10pt, conference, compsocconf]{IEEEtran}

\usepackage{stmaryrd}
\usepackage{amssymb}
\usepackage{amsmath}
\usepackage{color}
\usepackage{listings}
\usepackage{float}
\usepackage{booktabs}
\usepackage{threeparttable}
\usepackage{color, colortbl}
\usepackage{graphicx}
\usepackage{hyperref}
\usepackage{adjustbox}
\usepackage{multirow}
\usepackage{algorithm2e}
\usepackage{caption}
\usepackage{makecell}

\graphicspath{{./pdf/}{./jpeg/}{./jpg/}{./png/}}
\DeclareGraphicsExtensions{.pdf,.jpeg,.jpg,.png}
\usepackage{array}
\newcolumntype{C}{>{\centering\arraybackslash}p{1em}}

\definecolor{Gray}{gray}{0.95}

\usepackage{cite}
\usepackage{amsmath,amssymb,amsfonts}
\usepackage{algorithmic}

\IEEEoverridecommandlockouts    
    
\begin{document}

\title{ServeNet: A Deep Neural Network for \\Web Services Classification}

\author{
\IEEEauthorblockN{Yilong~Yang\IEEEauthorrefmark{3}, Nafees~Qamar\IEEEauthorrefmark{4}, Peng~Liu\IEEEauthorrefmark{3}, Katarina~Grolinger\IEEEauthorrefmark{5}, Weiru Wang\IEEEauthorrefmark{2}*, Zhi~Li\IEEEauthorrefmark{6}, Zhifang~Liao\IEEEauthorrefmark{8}
}


\IEEEauthorblockA{\IEEEauthorrefmark{3}Department of Computer and Information Science, \\ Faculty of Science and Technology,  University of Macau, Macau} 
\IEEEauthorblockA{\IEEEauthorrefmark{4}Governors State University, Chicago, United States}
\IEEEauthorblockA{\IEEEauthorrefmark{5}Department of Electrical and Computer Engineering, \\ University of Western, Ontario, Canada}
\IEEEauthorblockA{\IEEEauthorrefmark{6}College of Computer Science and Information Technology, Guangxi Normal University, Guilin, China}
\IEEEauthorblockA{\IEEEauthorrefmark{8}School of Computer Science and Engineering, Central South University, Changsha, China}
Email: yylonly@gmail.com (Yilong Yang)

%
\thanks{*Corresponding author: Weiru Wang (belinda.wwr@connect.um.edu.mo)}}





\maketitle

\begin{abstract}
Automated service classification plays a crucial role in service discovery, selection, and composition. Machine learning has been widely used for service classification in recent years. However, the performance of conventional machine learning methods highly depends on the quality of manual feature engineering. In this paper, we present a novel deep neural network 
to automatically abstract low-level representation of  {\color{black}both service name and service description} to high-level {\color{black}merged} features without feature engineering {\color{black}and the length limitation}, and then predict service classification on 50 service categories. To demonstrate the effectiveness of our approach, we conduct a comprehensive experimental study by comparing 10 machine learning methods on 10,000 real-world web services. The result shows that the proposed deep neural network can achieve higher accuracy in classification and more robust than other machine learning methods.
\end{abstract}


\section{Introduction}

Software reuse is treated as a promising way to reduce the cost of software development since last decades. Web services provide a unified and loosely-coupled integration to reuse the heterogeneous software components \cite{yang2018microshare}. Notwithstanding the advancement of the service and cloud computing in recent years, more and more high quality and reliable web services are available in public repositories, which are the valuable resources for software reuse \cite{Yang2018}. Two popular web services repositories are the standard Universal Description, Discovery, and Integration (UDDI) registry \cite{DBLP:conf/icws/Mintchev08} and APIs sharing platform. These public repositories allow service providers to publish services with their specification, which includes service description in natural language, service {\color{black}name}, URL, and search keywords (labels or tags). The key to software reuse is to find the required services in the repositories to satisfy the requirements based on the service specifications, which is the primary concern in service discovery \cite{DBLP:conf/icws/HajlaouiOBB17}. 

Two search approaches are widely used in service discovery \cite{DBLP:conf/icsoc/Hasselmeyer05}\cite{DBLP:conf/icws/ChengZLC16}. The first one is the keyword-based method. Service consumers use several keywords to search the candidate services first, and then select the target service based on descriptions. The second one is semantic search based on the semantic web services such as WDSL-S, OWL-S, WSMO \cite{DBLP:conf/icws/ChengZLC16}. The target services are matched according to the signatures of the service, i.e., the input and output parameters of the services, and then validated by the formal specification of the services \cite{DBLP:journals/jwsr/YangYLW14}. Although semantic search methods have higher accuracy than keyword-based methods, they only work when we provide the semantic information of all the services in the repository as well as the semantic information in the service query. That could be too ideal to be practical in a real-world situation. Moreover, semantic search methods are less efficient than keyword-based methods. Therefore, keyword-based methods are the first choice of most cases in practice.

The success of keyword-based searching highly relies on the quality of service keywords, which are manually assigned by developers. However, the assigned keywords are not always reliable and adequate. This is mainly because the developers may have the difficulty in choosing the best keywords from a large candidate pool, and lack the knowledge of all candidates. The limitation of manual keyword assignment creates the need for automated keyword prediction and tag recommendation through machine learning methods.

In this paper, we present a novel end-to-end deep neural network \textit{ServeNet} for web service classification. 
It can automatically abstract low-level representations from service name and service description to high-level features respectively and can merge the high-level features to unified features for service classification.
The experiment results demonstrate that the proposed \textit{ServeNet} achieves higher top-5 accuracy and top-1 accuracy than other machine learning methods. 

\vspace{.1cm}
\noindent The contributions of this paper are summarized as follows: 
\begin{itemize}

    \item A novel end-to-end deep neural network \textit{ServeNet} is proposed. It can not only automatically abstract the features from service description like the original \textit{ServeNet} \cite{YANG-ICWS-19} but also the features from service name, and then merge them into unified features without feature engineering and {\color{black}the length limitation of service name and description.}

    
    \item The proposed \textit{ServeNet} solves the sparse and context-independent issues in the original \textit{ServeNet} by {\color{black}introducing the dynamic embedding on mini-batch sets}.
    
    \item We demonstrate that the new proposed \textit{ServeNet} archive the higher accuracy than other machine learning methods for service classification on 50-category benchmark.
    
        
    
\end{itemize}

The remainder of the paper is organized as follows: Section 2 introduces the related work and the details of comparison between the original \textit{ServeNet} and the new proposed \textit{ServeNet}. Section 3 provides the architecture of \textit{ServeNet}. Section 4 presents the evaluation of \textit{ServeNet} and the benchmark with other machine learning methods. Section 5 concludes the paper and proposes the future work.

\section{Related work}
Service classification and service tag recommendation are to predict the keywords or categories of web services. We treat them as the same problem in this paper. We review the conventional machine learning methods first and then compare the state-of-the-art deep learning methods with the new proposed approach for service classification. 

\subsection{Conventional Machine Learning Methods}
{\color{black}There are several works using conventional machine learning methods to automatically predict the keywords (tags) of services. {\color{black}The work \cite{FANG12} clusters services using Web Services Description Language (WSDL) documents and predicts (enrichs) the tags according on the tags of other services in the same cluster.} 
The work \cite{ioannis2009}\cite{hongbing2010} compared several machine learning methods such as Naive Bayes \cite{DBLP:conf/icml/SuSM11}, Support Vector Machines (SVM) {\color{black}\cite{zhang18}}, k-Nearest Neighbors (kNN) \cite{DBLP:journals/tkde/AlmalawiFTCK16} and C4.5 \cite{DBLP:conf/aaai/MaliahS18} for service classification on 7 categories. They show that SVM has the best accuracy than other machine learning methods. 
{\color{black}The work \cite{Kapitsaki2014} adopts ensemble schemes by combining the classifiers among Naive Bayes, KNN, and C4.5 based on service description and WSDL.}
By integrating Latent Dirichlet Allocation (LDA) \cite{DBLP:conf/icml/CongCLZ17} with SVM for feature extractions, the work \cite{xumin2016} shows that LDA-SVM model can reach around 90\% accuracy on 10 categories. 
However, when predicting web services that contains multi-labels,
only less than 50\% accuracy can be achieved \cite{tingting2016}\cite{weishi2017}.

The current work of service classification heavily relies on the quality of feature engineering. Feature engineering takes advantage of human ingenuity and prior knowledge to compensate for the weakness of inability to extract and organize the discriminative information from the data \cite{bengio2013}, which is normally both difficult and expensive.

\subsection{Deep Learning}
Deep learning is provided as a promising alternative which can automatically abstract low-level representation from raw data to high-level features without feature engineering \cite{lecun2015deep}. It has been successfully applied to many fields such as image and text classification even with a large number of categories. For example, deep Convolutional Neural Network (CNN) \cite{DBLP:conf/cvpr/BagherinezhadRF17} can achieve less than 5\% top-5 errors of 1000 categories on ImageNet dataset \cite{DBLP:conf/cvpr/DengDSLL009}. For text classification problems, deep neural networks have achieved great success as well. The Recurrent Neural Networks (RNN) and Long Short-Term Memory (LSTM) have been widely used in text processing, they have the good performance for learning and processing the representation of text \cite{pmlr-v48-johnson16}. CNN has also been applied to text classification problems \cite{DBLP:conf/ijcai/WangHD18}. Especially when integrating with the recurrent model, recurrent-CNN can achieve better performance \cite{DBLP:conf/naacl/LeeD16}. Recurrent-CNN proposed a new context-based RNN with one-dimensional (1-D) convolution layer \cite{DBLP:conf/aaai/LaiXLZ15}. C-LSTM \cite{DBLP:journals/eswa/KimC18} stacks the one-dimensional (1-D) convolution layer with LSTM layer. In our previous work \cite{YANG-ICWS-19}, \textit{ServeNet} stacks 2-D CNN with Bi-directional LSTM (Bi-LSTM) for automated feature extraction. The 2-D CNN can extract not only local features between adjacent words but also the small regions (word stem) inside of words than 1-D CNN. Bi-LSTM can extract sequential features from both past and future. The stacked 2-D CNN and Bi-LSTM makes \textit{ServeNet} achieve better performance on the benchmark.
In short, all of the stacked convolutional and recurrent neural networks can achieve the best accuracy of classification on their benchmarks, because they can extract not only local features between adjacent words through convolution layers but also global features (long-term dependencies) of sentences by recurrent layers. 
\begin{table}[!htb]
  \centering
    \caption{{\color{black}ServeNet (Original) vs ServeNet}}
    \label{comparison}
 \begin{adjustbox}{max width=0.48\textwidth}
 \renewcommand{\arraystretch}{1.115}
  \begin{tabular}{c>{\centering}m{0.35\linewidth}>{\centering\arraybackslash}m{0.35\linewidth}}
    \toprule 
      & ServeNet (Original) & ServeNet \\
    \midrule
    \rowcolor{Gray}
    Input Parameter &   Service Description & \makecell{Service Name \\ Service Description}\\
    
    Input Length  & $\le$ 110 words & Arbitrary \\
    \rowcolor{Gray}
    Embedding Method  & Fixed Embedding
(GloVe) 
 & Context-aware Embedding
(BERT)
 \\

    Padding & \makecell{Fixed-length Padding \\ on all services}  & \makecell{Dynamic Padding \\ on Mini-batch} \\  \rowcolor{Gray}
    Dataset  & Remove Services (Description $>$ 110 words) & All Services\\
    
    Accuracy  & 88.40\%  & 91.13\%\\
    \bottomrule
  \end{tabular}
  
  
\end{adjustbox}
\end{table}

{\color{black} The related deep learning methods and original \textit{ServeNet} in our previous work have serval problems. 1) They only use service description to classify web services, i.e., they do not use any other information in service specification for classification task. 2) They use context-independent embedding \cite{8029857} (e.g. Word2Vec and GloVe) to encode the web services, that makes each word of service description embedded as a fixed vector without considering the position in a sentence. 3) They use zero batch padding to pad all the descriptions of web services to the same length, it will lead dataset sparse when the dataset contains a few long sentences. To alleviate the imbalance problem, the input length of service description is limited to 110 words and any service is removed from dataset when the length of the service description is above 110 in the original \textit{ServeNet}.
 
The new proposed \textit{ServeNet} solves above problems. 1) It can not only abstract the features from service description but also the features from service name, and then automatically merge them into unified features. 2) It adopts Bidirectional Encoder Representations from Transformers (BERT) \cite{devlin-etal-2019-bert} for dynamic context-dependent word embedding of on both service name and service description. BERT is a context-dependent word embedding method, which can generate different word embedding for a word that captures the context of a word - that is its position in a sentence. 3) The easier but inefficient way to make model accept arbitrary length input is set (\textit{BatchSize = 1}) in both training and testing phases, because LSTM and BERT models can accept arbitrary time steps for inputs. To achieve high efficiency in training, we use mini-batch padding to pad service description and name to the same length when the services are in one mini-batch. In meanwhile, \textit{ServeNet} computes the embedding mask for the inputs, which can indicate the valid information of sentences. The summary of the comparison between the original \textit{ServeNet} and the new proposed \textit{ServeNet} is shown in Table \ref{comparison}.

In short, the new proposed \textit{ServeNet} can automatically extract features separately from service name and description in arbitrary length, then merge them as unified features, and finally predict the service category. This new design architecture makes \textit{ServeNet} archive the highest accuracy on the benchmark.
}



\begin{figure*}[!htb]
    \centering
    \includegraphics[width=1\textwidth]{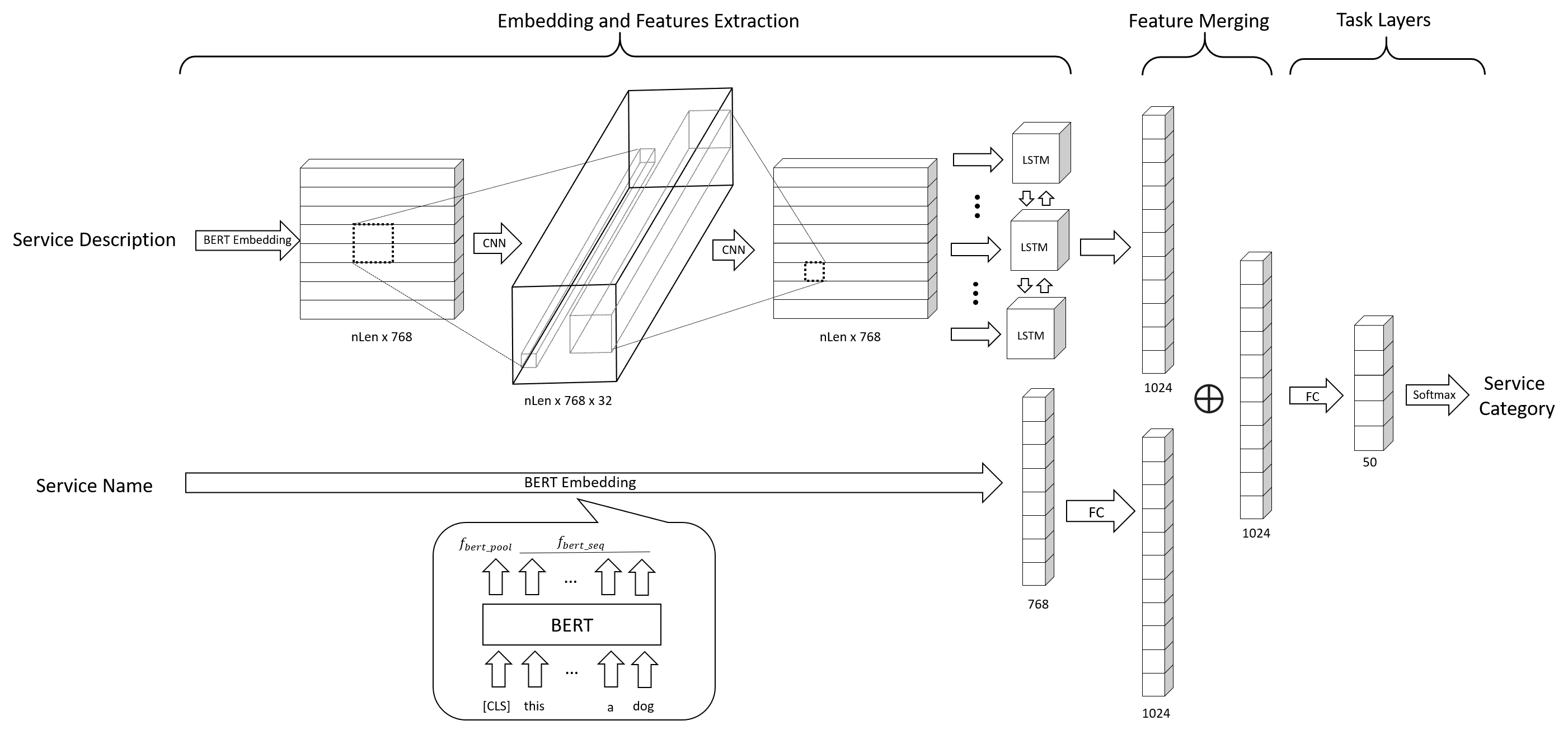}
    \caption{ServeNet architecture}
    \label{ServeNet Architecture}
\end{figure*}

\section{ServeNet}
The architecture design of a neural network is the key to apply deep learning to domain-specific problems. In this section, we introduce the architecture of the proposed deep neural network \textit{ServeNet} as well as its hyper-parameters.

\subsection{ServeNet Architecture}
The proposed deep neural network \textit{ServeNet} contains four parts: a) service description embedding and its feature extraction, b) service name embedding and its feature extraction, c) feature merging, and d) task layers. The structure overview of \textit{ServeNet} is shown in Fig. \ref{ServeNet Architecture}.

\subsubsection{Service description embedding and feature extraction}

The architecture of this part is almost same as the original \textit{ServeNet} \cite{YANG-ICWS-19} except using 
BERT model as the embedding layer, which is a pre-trained language model and can transform a word of description to a $n$-dimensional vector according to the contexts of words. 
{\color{black}In addition, we use dynamic padding to each mini-batch set for accepting arbitrary length of inputs. The services of each mini-batch set are dynamically padded to the same length \textit{nLen} (\textit{nLen} is the max length in the mini-batch). Note that by providing a sentence, BERT can output two types of embedding: a) $f_{bert\_pool}$ embeds a sentence to a vector, it is used to embed a short sentence, in which no further feature extraction are required and b) $f_{bert\_seq}$ embeds each word of a sentence to a vector, which outputs a embedding matrix. This matrix keeps the order and representation of each word, further feature extraction can be done based on this matrix if needed. In this paper, we use $f_{bert\_pool}$ to embed service name and $f_{bert\_seq}$ to embed service description.}

The embedding layer of service description outputs a \textit{nLen} by \textit{n} description matrix $e$ for the next layer. We assume that $x_1$ is the description of a service. The embedding of service description $f_{bert\_seq}$ can be defined as:
\[
    e = f_{bert\_seq}(x_1)
\]
Same as the original \textit{ServeNet}, the new \textit{ServeNet} also adopts the stacked 2-D CNN and Bi-LSTM for feature extraction of service description. Therefore, the output feature $h_1$ of service description is:
\[
    h_1 = f_{lstm} \cdot  f_{cnn} (e)
\]

\subsubsection{Service name embedding and feature extraction}

BERT not only can output a sequential embedding matrix but also can output an embedding vector for a sentence. The service name usually contains a few words, the further complex feature extraction are not necessary. After getting the vector embedding of service name from BERT output, we use a fully connected neural network $f_{\textit{full}}$ with activation function $tanh$ for feature extraction. We assume that $x_2$ is the name of a service. The embedding and feature extraction of a service name can be defined as:
\[
    h_2 =  f_{\textit{full}} \cdot f_{bert\_pool}(x_2)
\]

\subsubsection{Feature Merging}
After retrieving the feature vectors of service description $h_1$ and service name $h_2$, we merge them into a unified feature. The lengths of those features are same in the proposed architecture, we adopt simple vector addition for feature merging. The output of feature merging is:
\[
    h = h_1 + h_2
\]

\subsubsection{Task layers}
The task layers do the final classification task. It contains a fully connected feed-forward neural network $f_{fc}$, which inputs high-level representation $h$ from feature extraction layers and outputs a service classification $l$:
\[
    l = f_{fc}(h)
\]
$f_{fc}$ contains two fully connected layers with activation function $\sigma_1$ and $\sigma_2$. Each layer computes $a_{i+1}$ by the weight $W_{i}$, bias $b_i$, and the output $a_{i}$ from the previous layer:
\[ 
a_{i+1} = \sigma (W_{i}*a_{i} + b_i)
\]
where $\sigma_1$ is the \textit{tanh} function and $\sigma_2$ is the \textit{softmax} function that computes the probability of each category. 

In short, \textit{ServeNet} takes service description $x_1$ and service name $x_2$ as input, and outputs service classification $l$: 
\[l = f_{fc} \cdot ( f_{lstm} \cdot f_{cnn} \cdot  f_{bert\_seq}(x_1) + f_{\textit{full}} \cdot f_{bert\_pool}(x_2))\]


\subsection{Hyper-parameters}
The hyper-parameters of \textit{ServeNet} contains the network configuration and training setting. 

\subsubsection{Network hyper-parameters} 
We choose the pre-trained BERT model (\textit{bert\_en\_uncased\_L-12\_H-768\_A-12}) from Tensorflow Hub\footnote{\url{https://tfhub.dev/tensorflow}}, which transforms each word of service description and service name into a 768-dimension vector. \textit{ServeNet} contains two convolution layers and one bidirectional LSTM layer. The first convolution layer has 32 filters with kernel size 3 by 3, the second convolution layer has 1 filter with kernel size 1 by 1. The hidden state of LSTM is a 1024-dimension vector. The task layer contains 50 nodes with an activation function \textit{softmax} to compute the probabilities of each category. To avoid over-fitting, we add a dropout layer between every two layers of \textit{ServeNet} with drop probability (rate) 0.1.

\subsubsection{Training hyper-parameters} 
\textit{ServeNet} adopts the categorical cross-entropy as the loss function. The Adam optimization algorithm is used for training with the learning rate 2e-5, beta1 0.9, beta2 0.999, epsilon=1e-6, and learning decay 0.01. The total epoch number is 20 with batch size 64. The hidden state of LSTM and all bias are initialized to zero. Xavier normal initializer is used to initialize the kernel parameters except the task layer initialized by Truncated normal initializer with the standard deviation 0.02.

\section{Evaluation}
In this section, we introduce the service dataset, evaluation metrics, and the evaluation results of \textit{ServeNet}, which includes the comparison results with other machine learning methods.

\subsection{Service Dataset}

\subsubsection{Service Collection}

Web services are collected from API sharing platform\footnote{\url{http://www.programmableweb.com}}, which includes all types of service such as SOAP, REST, RPC, and gRPC. {\color{black}Over 84\% services are REST style. Less 8\% services are in SOAP style, which makes the additional WSDL hard to be retrieved for service classification.} Service specification provided by API sharing platform includes \emph{title}, \emph{description}, \emph{endpoint}, \emph{homepage}, \emph{primary category}, \emph{secondary categories}, \emph{provider}, \emph{SSL support}, and etc. We implement a web crawler through web browser automation tool Selenium\footnote{\url{https://www.seleniumhq.org}} to collect services and store them into a service dataset \textit{WSDataset} in CSV format.




The original \textit{WSDataset} contains 15344 services. We clean the dataset to exclude the services of which the title, descriptions or catalogs are empty. After this basic cleanup, the dataset remains 15340 services with 401 categories, each service is specified by 20 descriptors. In this paper, we only take service descriptors - \textit{title},  \textit{description} and \textit{primary category} into account. To prevent ambiguity, we define \textit{title} as service name, \textit{description} as service description, \textit{primary category} as service classification. Note that service classification and category are exchangeable concepts in this paper. 



\subsubsection{Service Category Analysis}
After counting and ranking the number of services in each category, we found
the maximum category is \textit{Tools}, which contains 767 services, while the minimum categories are categories with only one service, which are one-shot categories. Moreover, only 41 categories contain the number of services greater than 100, and 217 categories (more than the half of categories) contain the number of services less than 10. In short, the service dataset is extreme imbalance. 
In this paper, we eliminate the one-shot, few-shot, small size categories and keep big size categories to make dataset more balance. 
Note that the top 50 categories include \textit{Other} category. We do not add the removed services into this category because it will make dataset imbalance. After this processing, service dataset contains 10943 services with 50 categories.

\subsubsection{Trainning and testing data selection}


Data selection is an important step of dataset pre-processing. There are several methods available, such as random selection and k-fold cross-validation. In order to apply the suitable method to service classification on 50 categories, we make a comparison experiment on a) 10-time random selection, b) 10-fold cross-validation, c) 10-time random selection by keeping the same percentage on each category, and d) 10-fold cross-validation by keeping the same percentage on each category through the classical text classification method Naive-Bayes on Top-5 accuracy Note that the explanation of Top-5 can be see in the next subsection.

\begin{table}[!htb]
  \centering
    \caption{{\color{black}Comparison of data selection methods}}\label{splitingmethod}
   \begin{adjustbox}{max width=0.48\textwidth}
 \begin{threeparttable}
  \begin{tabular}{ccc}
    \toprule
    Splitting method & Accuracy\_M & Accuracy\_V\\ 
    \midrule
    Random selection & 73.2363& 0.0189 \\
    10-fold cross-validation  & 74.6075  & 0.0359 \\
    Random selection by category & \textbf{75.6914} & \textbf{0.0035}\\
    10-fold cross-validation by category & 74.8000& 0.0150 \\    
    \bottomrule
  \end{tabular}
  
  \begin{tablenotes}[para]
  \footnotesize
         \item[*] Accuracy\_M stand for the mean values of the accuracies on testing set. Accuracy\_V stand for the standard deviation of the accuracies on testing set. The unit of the value is the percentage.
  \end{tablenotes}
  
  \end{threeparttable}
  \end{adjustbox}
\end{table}

{\color{black}We compute each mean value and standard deviation of the accuracy on the training set and the testing set in Table \ref{splitingmethod}. The result shows that the accuracies are close. Random selection by keeping the same percentage on each category reaches the highest accuracy and the lowest standard deviation on both training and testing sets. Because the service dataset is small and imbalanced, random selection cannot make the training set and test set conform to the same distribution on the small size categories. Even worse for some categories, the training set is smaller than the testing set. Random selection by category can keep the same proportion of training and testing data on each category, which can improve the accuracy of classification. Therefore, random selection by category is used to data selection. The training set contains 8733 services. and testing set includes 2210 service. We use them to train and test the proposed model and other machine leaning methods.}

{\color{black}\subsubsection{Dynamic Padding on Mini-batch} 
The original \textit{ServeNet} uses the fixed-length padding to pad all the descriptions of web services to the same length, it will lead dataset sparse because the length of description in service dataset is an imbalance. To alleviate this problem, the input length of the service description is limited to 110 words. All the service with service description ($>$ 110 words) will be removed from the dataset. However, this approach sacrifices the extensibility in practice. 
In this paper, we use dynamic padding to each mini-batch for accepting arbitrary length of inputs, because LSTM and BERT can accept arbitrary time steps in one batch. Compared with the easier but inefficient way to make model accept arbitrary length of inputs by setting \textit{BatchSize = 1} in both training and testing phases, we first sort service in terms of the length of the service description and then we embed the services in each mini-batch to the same length \textit{nLen}, which is the max length in a mini-batch set. In addition, we replace the input shape of \textit{ServeNet} from \textit{(maxLen, )} to \textit{(None, )} in TensorFlow source code. After these processing, our proposed \textit{ServeNet} can accept arbitrary length inputs without losing training efficiency.}

\subsection{Evaluation Metrics}
Top-N accuracy is used to evaluate a classification model. For example, Top-1 accuracy is usually used to evaluate a binary classification model. A multi-classes classification model (classes $>$ 10, e.g., ImageNet) uses Top-5 evaluation metrics. When providing a service with name and description, \textit{ServeNet} predicts the probability of each category in vector $l_{pre}$. Top-5 prediction is correct when Top-5 predicted labels contain the target category $l_{\textit{target}}$. Top-5 accuracy on the category $c$ is :

\[
\textit{Accuracy}_{top5}(c) = \frac{\textit{Num}_{top5\_\textit{correct}}}{\textit{Num}_c} 
\]
where $\textit{Num}_{top5\_\textit{correct}}$ is the number of correct prediction, $\textit{Num}_c$ is the number of the service in category $c$. Top-5 accuracy on the whole dataset is :
\[
\textit{Accuracy}_{top5} = \sum_{c \; \in \; \textit{categories}} \frac{\textit{Accuracy}_{top5}(c)}{\textit{Num}_{\textit{categories}}}
\]
where $\textit{Num}_{\textit{categories}}$ is the number of the categories in the dataset. Top-1 accuracy is similar to Top-5 accuracy, except the correct prediction means the Top-1 prediction is the same as the target category $l_{target}$.

\subsection{Experiment Result and Discussion}
{\color{black}We compare 10 machine learning methods from conventional machine learning methods to deep learning models for service classification on the proposed service dataset, which includes Naive-Bayes, Random Forest (RF) \cite{pmlr-v80-haghiri18a}, SVM (LDA-Linear-SVM and LDA-RBF-SVM), AdaBoost, CNN, Recurrent-CNN, LSTM, BI-LSTM, and C-LSTM.} {\color{black} All the conventional models are implemented by scikit-learn library\footnote{\url{https://scikit-learn.org}} and all the deep learning models are implemented by TensorFlow.}
The source codes and dataset are available on GitHub\footnote{\url{https://github.com/yylonly/ServeNet}}. 
{\color{black}All the models are trained and tested on the Dell T630 server with Nvidia GTX1070 and Google Colab\footnote{\url{https://colab.research.google.com}}.}

\begin{table}[!htb]
  \centering
    \caption{{\color{black}Comparison result of machine learning methods}}\label{meanACC}
 \begin{adjustbox}{max width=0.5\textwidth}
  \begin{tabular}{ccc}
    \toprule
    Model & Top-5 Accuracy & Top-1 Accuracy \\
    \midrule
    CNN &   58.46 & 27.60\\
    AdaBoost  & 64.92 & 34.93 \\
    LDA-Linear-SVM  & 71.91 & 33.28 \\
    LDA-RBF-SVM & 73.84  & 39.79\\
    Naive-Bayes  & 78.94 & 47.74\\
    LSTM  & 80.10  & 51.18\\
    RF  & 80.25 & 54.29\\
    Recurrent-CNN & 84.29 & 60.02\\
    C-LSTM  & 84.32 & 59.24\\
    BI-LSTM &  86.70 & 60.45\\
    ServeNet (Orignal)  & 88.40  & 63.31\\
    ServeNet  & \textbf{91.58} & \textbf{69.95}\\
    \bottomrule
  \end{tabular}
  
  
\end{adjustbox}
\end{table}

\subsubsection{Experiment Result}
{\color{black}The experiment results are shown in Table \ref{meanACC}. Note that the rank of Top-1 results is similar to Top-5, we focus on discussing Top-5 accuracy in this paper. CNN has the lowest accuracy 58.46\%, which means that only considering the local features between adjacent words is not working for service classification problem. {\color{black}The new proposed \textit{ServeNet} can achieve the highest test accuracy on both Top-5 accuracy 91.58\% and top-1 accuracy 69.95\% because it can automatically extract features from both service name and description with context-dependent embedding.}

\begin{table*}[!htb]
  \centering
    \caption{{\color{black}Comparison results of top-5 accuracy on each category}}\label{top-5all} 
  \begin{adjustbox}{max width=\textwidth}
  \begin{tabular}{ccccccccccccc}
  \toprule
Service Category        & CNN             & AdaBoost & LDA-Linear-SVM & LDA-RBF-SVM    & Naive-Bayes      & LSTM            & RF              & Recurrent-CNN   & C-LSTM          & BI-LSTM         & ServeNet(Original) & ServeNet        \\
    \midrule
Tools                   & 71.23           & 97.95    & 91.10          & 95.20          & 99.32           & 85.62           & \textbf{100.00} & 89.04           & 82.88                    & 91.78           & 89.04    & 83.78      \\
Financial               & 84.62           & 95.38    & 76.15          & 76.94          & 97.69           & 93.85           & \textbf{100.00} & 93.08           & 93.85                     & 97.69           & 97.69     &  90.41   \\
Messaging               & 85.57           & 83.51    & 92.78          & 95.91          & 96.91           & 92.78           & \textbf{96.91}  & 95.88           & 95.88                     & 93.81           & 94.85     &    93.33    \\
eCommerce               & \textbf{100.00} & 88.37    & 82.56          & 88.41          & 94.19           & \textbf{100.00} & 98.84           & \textbf{100.00} & \textbf{100.00} &   \textbf{100.00} & \textbf{100.00} & \textbf{100.00}\\
Payments                & 87.06           & 80.00    & 85.88          & 90.60          & 96.47           & 88.24           & 91.76           & \textbf{97.65}  & 95.29                    & 92.94           & 96.47         &   95.15\\
Social                  & 48.75           & 87.50    & 77.50          & 90.00          & \textbf{97.50}  & 83.75           & 96.25           & 88.75           & 81.25                      & 88.75           & 90.00       & 96.67    \\
Enterprise              & 55.70           & 75.95    & 62.03          & 65.80          & \textbf{96.20}  & 63.29           & 94.94           & 69.62           & 74.68                      & 81.01           & 87.34         & 90.70  \\
Mapping                 & 59.70           & 82.09    & 82.09          & 86.61          & 97.01           & 86.57           & 92.54           & 83.58           & 92.54                    & 91.04           & \textbf{97.01}  & 95.45\\
Telephony               & 54.39           & 82.46    & 80.70          & 63.22          & \textbf{100.00} & 92.98           & 84.21           & 96.49           & 89.47                     & 84.21           & 92.98       & 94.87    \\
Science                 & 78.18           & 74.55    & 72.73          & 72.72          & 89.09           & 94.55           & 89.09           & \textbf{98.18}  & 96.36                      & 92.73           & \textbf{98.18}  & 80.00\\
Government              & 83.64           & 81.82    & 78.18          & 72.71          & \textbf{94.55}  & 89.09           & 74.55           & \textbf{94.55}  & 92.73                      & 92.73           & 89.09       & 89.66    \\
Email                   & 68.75           & 70.83    & 70.83          & 83.32          & 97.92  & 93.75           & 91.67           & 93.75           & 91.67                      & 91.67           & 91.67       & \textbf{98.04}    \\
Security                & 36.17           & 72.34    & 59.57          & 61.72          & 68.09           & 74.47           & 59.57           & 82.98           & 72.34                      & 85.11           & 87.23 & \textbf{94.12} \\
Reference               & 25.53           & 80.85    & 65.96          & 66.00          & 78.72           & 63.83           & 36.17           & 55.32           & 59.57                      & 53.19           & 82.98 & \textbf{89.47} \\
Video                   & 74.47           & 89.36    & 87.23          & 89.41          & 97.87           & 97.87           & 97.87           & \textbf{100.00} & 97.87            & 97.87           & 97.87           & 70.59\\
Travel                  & 62.22           & 71.11    & 77.78          & 91.12          & 86.67           & 88.89           & 86.67           & 95.56           & \textbf{100.00}            & 97.78           & 88.89       & 97.56    \\
Sports                  & 74.42           & 69.77    & 74.42          & 60.50          & 79.07           & 93.02           & 79.07           & 95.35           & 97.67             & 95.35           & 95.35      & \textbf{100.00}     \\
Search                  & 18.60           & 69.77    & 62.79          & 69.83          & 79.07           & 58.14           & 69.77           & 60.47           & 58.14                      & 74.42           & \textbf{83.72} & 72.22 \\
Advertising             & 54.76           & 76.19    & 76.19          & 78.62          & 83.33           & 83.33           & 80.95           &  90.48           & 88.10             & 88.10           & 76.19       & \textbf{100.00}    \\
Transportation          & 76.19           & 73.81    & 71.43          & 64.32          & 85.71           & 83.33           & 76.19           & 92.86           & \textbf{100.00} &  97.62           & 92.86           & 55.17\\
Education               & 43.90           & 58.54    & 60.98          & 63.42          & 60.98           & 85.37           & 82.93           & 90.24           & 90.24                     & 87.80           & 95.12  & \textbf{95.45}\\
Games                   & 48.72           & 64.10    & 71.79          & 74.43          & 87.18           & 76.92           & 89.74           & 94.87  & 87.18                     & 82.05           & 89.74      & \textbf{100.00}     \\
Music                   & 64.86           & 72.97    & 72.97          & 67.62          & 86.49           & 86.49           & 91.89           & 94.59           & \textbf{100.00}            & 97.30           & 91.89    & \textbf{100.00}       \\
Photos                  & 57.14           & 54.29    & 68.57          & 80.00          & 82.86           & 85.71           & 88.57           & 91.43           & 94.29                      & 94.29           & \textbf{97.14}  & 91.30\\
Cloud                   & 39.39           & 72.73    & 45.45          & 48.51          & 66.67           & 81.82           & 87.88  & 84.85           & 78.79                      & 81.82           & 84.85       &  \textbf{90.00}  \\
Bitcoin                 & 78.57           & 89.29    & 82.14          & 82.12          & 85.71           & \textbf{100.00} & 96.43           & 96.43           & 89.29                      & \textbf{100.00} & 92.86        & 96.30   \\
Project Management      & 32.14           & 53.57    & 64.29          & 53.62          & 82.14           & 64.29           & 71.43           & 75.00           & 82.14                     & 64.29           & \textbf{85.71} & 82.61 \\
Data                    & 35.71           & 17.86    & 25.00          & 28.62          & 17.86           & 42.86           & 10.71           & 42.86           & 50.00                      & 42.86           & 64.29 & \textbf{90.91} \\
Backend                 & 37.04           & 25.93    & 37.04          & 59.31          & 40.74           & 48.15           & 29.63           & 70.37           & 70.37                      & 74.07  & 66.67      & \textbf{85.19}     \\
Database                & 22.22           & 33.33    & 25.93          & 29.63          & 14.81           & 37.04           & 18.52           & 44.44           & 44.44                      & 62.96  & 59.26      & \textbf{96.81}     \\
Shipping                & 69.23           & 69.23    & 80.77          & 76.93          & 80.77           & 84.62           & 96.15           & \textbf{96.15}  & 96.15                      & \textbf{96.15}  & \textbf{96.15} & 84.21 \\
Weather                 & 56.52           & 91.30    & 82.61          & 87.00          & 86.96           & 95.65           & 95.65           & 91.30  & 91.30                      & 91.30  & 91.30      & \textbf{100.00}     \\
Application Development & 21.74           & 21.74    & 39.13          & 21.71          & 13.04           & 56.52           & 8.70            & 69.57           & 78.26                      & 82.61  & 82.61 & \textbf{93.10} \\
Analytics               & 26.09           & 43.48    & 43.48          & 52.21          & 13.04           & 43.48           & 30.43           & 60.87           & 39.13                      & \textbf{91.30}  & 82.61     &  62.50      \\
Internet of Things      & 13.64           & 54.55    & 54.55          & 54.52          & 59.09           & 68.18           & 54.55           & 54.55           & 72.73            & 36.36           & 59.09     & \textbf{92.77}       \\
Medical                 & 23.81           & 52.38    & 57.14          & 47.62          & 19.05           & 95.24           & 31.25           & \textbf{100.00} & 90.48                      & 85.71           & \textbf{100.00} & 94.12\\
Real Estate             & 42.86           & 52.38    & 71.43          & 90.53 & 61.90           & 76.19           & 85.71           & 80.95           & 80.95                     & 76.19           & 90.48    & \textbf{96.08}       \\
Events                  & 23.81           & 71.43    & 95.24 & 95.21          & 52.38           & 71.43           & 85.71           & 90.48           & 85.71             & 85.71           & 95.24 & \textbf{98.04} \\
Banking                 & 80.00           & 45.00    & 70.00          & 85.00          & 65.00           & 95.00           & 80.00           & 90.00           & 80.00                      & \textbf{100.00} & 85.00     & 87.66      \\
Stocks                  & 89.47           & 78.95    & 94.74          & 94.72          & 89.47           & \textbf{100.00} & 94.74           & \textbf{100.00} & 94.74                     & \textbf{100.00} & 94.74       & 95.00    \\
Entertainment           & 10.53           & 15.79    & 31.58          & 26.31          & 21.05           & 63.16           & 21.05           & 63.16           & \textbf{73.68}             & 68.42           & 73.68  & 64.29\\
Storage                 & 21.05           & 57.89    & 73.68          & 63.22          & 31.58           & 68.42           & 73.68           & 78.95  & 78.95             & 78.95  & 73.68      & \textbf{91.95}     \\
Marketing               & 37.50           & 43.75    & 50.00          & 62.50          & 18.75           & 81.25           & 75.00           & 87.50  & 68.75                      & 81.25           & 75.00      & \textbf{95.45}     \\
File Sharing            & 50.00           & 62.50    & 68.75          & 43.82          & 56.25           & 62.50           & 68.75           & 81.25           & 75.00             & \textbf{87.50}  & \textbf{87.50} & 86.27 \\
News Services           & 25.00           & 37.50    & 62.50          & 68.80          & 37.50           & 56.25           & 50.00           & 75.00           & 81.25                      & 87.50  & 81.25     & \textbf{98.33}      \\
Domains                 & 56.25           & 81.25    & 87.50          & 81.20          & 75.00           & 75.00           & 75.00           & 87.50           & 93.75   & 87.50           & 87.50           & \textbf{96.42}\\
Chat                    & 68.75           & 37.50    & 93.75          & 75.00          & 62.50           & 81.25           & 68.75           & \textbf{100.00} & 93.75                      & 93.75           & 93.75       & 84.31    \\
Media                   & 31.25           & 31.25    & 37.50          & 37.50          & 6.25            & 50.00           & 31.25           & 68.75           & 81.25                      & 87.50  & 75.00    & \textbf{94.16}       \\
Images                  & 26.67           & 33.33    & 40.00          & 53.30          & 20.00           &  53.33           & 46.67           &  53.33           &  53.33            &  53.33           &  53.33      & \textbf{88.24}     \\
Other                   & 55.17           & 17.24    & 24.14          & 31.00          & 0.00            & 27.59           & 3.45            & 37.93           & 31.03                      & 62.07  & 62.07 & \textbf{88.57}\\
 \midrule
 $\sigma$ & 23.50 & 22.45 & 19.35 & 20.09 & 30.96 & 18.34 & 27.98 & 16.86 & 16.60 & 14.89 & 11.69 & \textbf{10.00} \\
   \bottomrule
  \end{tabular}
  \end{adjustbox}
 
\end{table*}

For the conventional methods, Naive-Bayes is a simple and well-performing model for text classification since the last decade. It can achieve the median top-5 accuracy 78.94\% among all models. By integrating LDA, SVM can achieve 71.91\% and 73.84\% accuracies on linear and RBF (Radial Basis Function) kernels respectively. We also benchmark the ensemble methods such as boosting and bagging. AdaBoost is a boosting method. It has 98\% accuracy on the training set, but only 64.92\% accuracy on the testing set, which shows around 30\% difference, that is a low bias but high variance model. AdaBoost can enhance a weak classifier to become a strong classifier continuously until the prediction accuracy of the training set reach the threshold. But it makes model easily over-fitting on the training set. Conversely, RF is a bagging method. It has 90\% accuracy on the training set and 80\% accuracy on the testing set. It has higher bias and lower variance than AdaBoost. This method helps to reduce the variance to alleviate over-fitting, because bagging uses bootstrap sampling to obtain the subsets of the dataset, then train the base learners from the subsets of the datasets, and finally aggregate the outputs of base learners by voting or averaging.

For the deep learning methods, sequence models abstract representation through time steps. They can learn global features such as long-term dependencies from the past time steps, which makes LSTM reach 80\% accuracy on the testing set. When stacking sequence model with 1-D CNN for learning the local features, both Recurrent-CNN (sequence model + CNN) and C-LSTM (CNN + sequence model) can get higher accuracy 84.29\% and 84.32\% with the improvement 4\%. If we compute time steps from both past and future, Bi-LSTM can learn the long-term dependencies, which makes Bi-LSTM reach 86.70\% accuracy. Finally, the original proposed \textit{ServeNet} uses 2-D CNNs with Bi-LSTM, it can learn both local and global features as well as the features in small regions inside of words (word stem). It can reach the higher accuracy 88.40\% and 63.31\% among all benchmarks}. {\color{black}Moreover, the new proposed \textit{ServeNet} can use both information from service name and description with context-dependent word embedding. It reaches the highest accuracy 91.58\% and 69.95\% among all benchmarks}.

}



{\color{black}
The benchmark of Table \ref{meanACC} only shows the average accuracy of 50 categories on each model. To show more details comparison of the proposed model, Top-5 accuracies of all categories are shown in Table \ref{top-5all}. 
We highlight the highest value in each category and compute the standard deviation of the categories for each machine learning method. Deep sequence models have lower standard deviation between each category and contain more categories with the highest accuracy than the conventional machine learning methods. It demonstrates that the deep sequence models are more robust than the conventional machine learning methods. Our proposed model \textit{ServeNet} has the most highlighted values and lowest variance among them. 

\begin{figure}[!htb]
    \centering
    \includegraphics[width=0.5\textwidth]{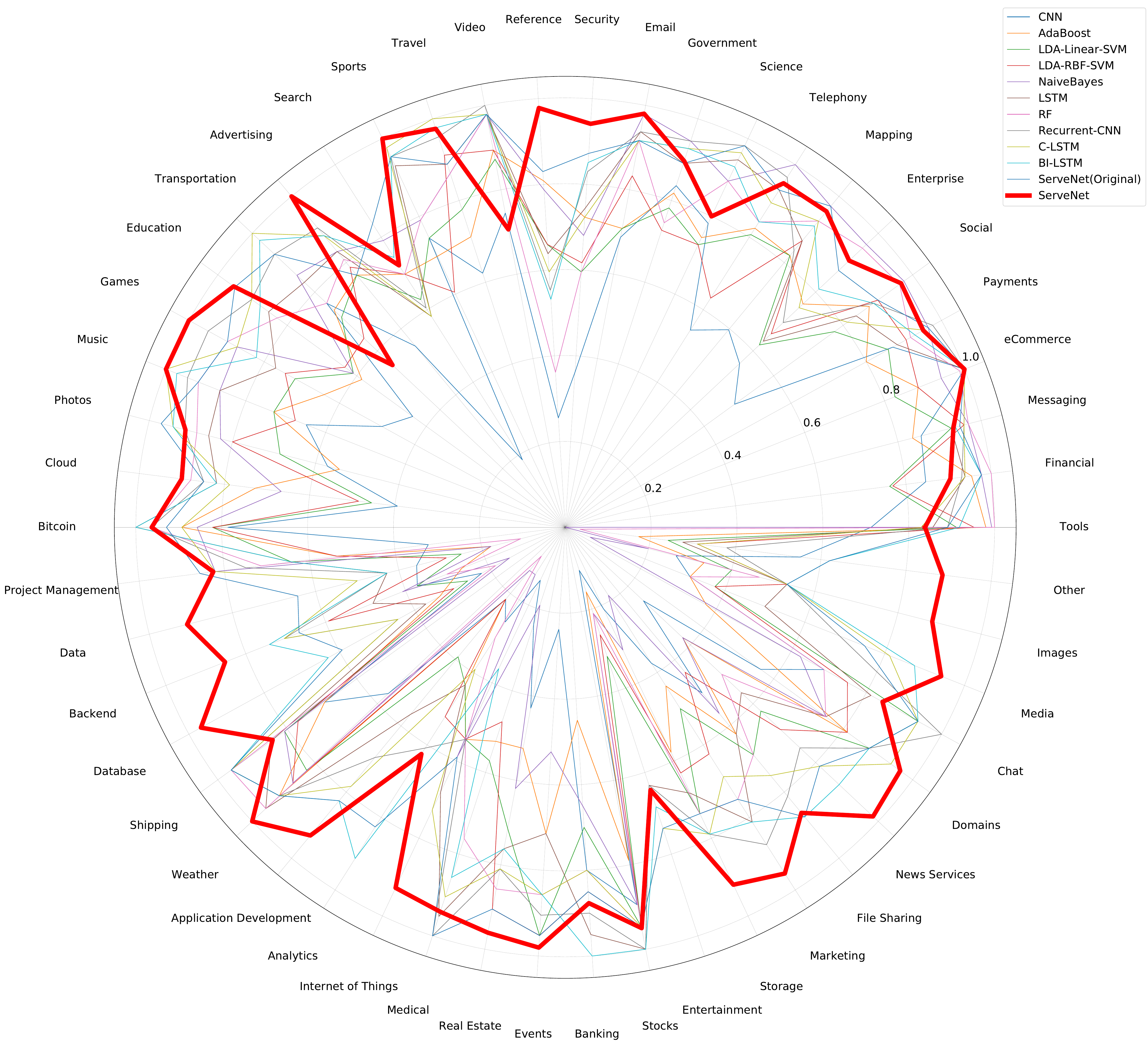}
    \caption{{\color{black}Top-5 accuracy on all categories}}
    \label{RadarTop5}
\end{figure}


After analyzing the name and description of services, we found if the categories contain clear topic features in the name and description such as \textit{Events}, LDA-based SVM has the good performance on those categories. If the categories contain a number of services such as \textit{Tools}, \textit{Financial}, and \textit{Messaging}, RF can reach the highest on those categories. Moreover, the categories such as \textit{Social}, \textit{Enterprise}, \textit{Telephony}, and \textit{Government} do not contain too many long-term dependencies in the descriptions, so Naive-Bayes works well. To compare the accuracy of all categories more intuitively, we show a radar chart in Fig. \ref{RadarTop5}. It displays a closed polygonal line for each category. The proposed \textit{ServeNet} locates in the most outside of the radar with the maximum area, which means it has the best performance over all other methods for service classification. 
}

\subsubsection{Limitation} {\color{black} \textit{ServeNet} has the highest accuracy and lowest variance in service classification problem of Fig. \ref{RadarTop5} and Table \ref{top-5all}, but it still does not work well on the categories with the limited number of data points. Besides,
we found that the failed predictions are mainly from the fact that those service name and descriptions do not contain the information about target categories but contain the information in other categories.} 
One possible way is to invent a more advanced model to enable the neural network to discover those conflicts. But the more efficient way is to collect more high-quality data in practice. {\color{black} Transfer learning is another a promising method for dealing with the small data problem between relevant domains. Although web services and Apps are different concepts of software development. In details, web services are reusable APIs in the level of software development. Apps are products for the end-users. App stores such as Apple and Android contains more data points with App name, description, and categories. We can pre-train a model on App dataset and then transfer it to the service classification problem, which will be a good direction to improve our work.}


\section{Conclusion and Future Work}
In this paper, we introduce a novel deep neural network \textit{ServeNet} to predict the service categories from service specifications. \textit{ServeNet} can automatically extract features from service name and description, then merge them into unified feature, and finally predict service classification. \textit{ServeNet} is trained and tested on 10943 services with 50 categories. The benchmark demonstrates that \textit{ServeNet} is robust and can achieve the best performance (Top-5 accuracy 91.58\% and Top-1 accuracy 69.95\%) than other machine learning methods. Although \textit{ServeNet} has higher prediction accuracy than other machine learning methods, there are still spaces to do more work to improve the model based on this paper. 

In the future, we consider to utilize more information from service signature such as the input and output parameters of services as well as other related deep learning models from application classification tasks to enhance this work. As noticed, the collected service dataset contains more than 400 categories, we consider to extend the task of services classification from 50 categories to 400 categories. In addition, we plan to integrate service classification with service refinement \cite{Yang19} to facilitate the development of web services. Moreover, our previous work present an approach \cite{YANG-TR-2019} and a CASE tool RM2PT \cite{DBLP:journals/corr/abs-1808-10657} to automatically generate prototypes from requirements models. Service discovery with automatic classification can enhance and tackle failures of the prototype generation by matching the non-executable requirements to a available web services. In addition, by integrating requirements validation through automatic prototyping \cite{YangRE19}, we can directly generate an implementation of web services from the validated requirements models. That will enhance the reliability and reduce overall cost of the development of web services. We believe our work can facilitate more research work in the related topics of service computing and other fields of software engineering.


\section{Acknowledgment}
This work was supported by the Ministry of Science and Technology: Key Research and Development Project (No. 201YFB003800) and National Natural Science Foundation of China (NSFC) (No. 61862009). In addition, the authors would like to thank B. Liu (Southwest University), Y. Zhao (East China Normal University), B. Yang (Norwegian University of Science and Technology) and B. Shen (University of Macau) for
proofreading the early version of this paper.

\bibliographystyle{IEEEtran}
\bibliography{service}

\end{document}